\documentclass[journal,10pt]{IEEEtran}
\usepackage[colorlinks,linkcolor=blue]{hyperref}
\usepackage{cite}

\usepackage{amsmath,amssymb,amsfonts}
\usepackage{graphicx}
\usepackage{textcomp}
\usepackage{subfigure}

\usepackage{booktabs}

\usepackage{multicol}
\usepackage{algpseudocode} 
\usepackage{algorithm}
\usepackage{caption}
\usepackage{multirow}
\usepackage{soul}
\usepackage{tikz}
\usepackage{lipsum,adjustbox}
\usepackage{amsmath}

\usetikzlibrary{positioning,chains}
\usetikzlibrary{shapes}
\usetikzlibrary{arrows}
\usetikzlibrary{positioning}
\usetikzlibrary{decorations.pathreplacing}

\usepackage[figurename=FIGURE.]{caption}

\def\BibTeX{{\rm B\kern-.05em{\sc i\kern-.025em b}\kern-.08em
		T\kern-.1667em\lower.7ex\hbox{E}\kern-.125emX}}

\begin{document}
\bibliographystyle{unsrt} 

\title{Hierarchical RNNs-Based Transformers MADDPG for Mixed Cooperative-Competitive Environments}

\DeclareRobustCommand*{\IEEEauthorrefmark}[1]{%
\raisebox{0pt}[0pt][0pt]{\textsuperscript{\footnotesize\ensuremath{#1}}}}

\author{
	Xiaolong Wei\IEEEauthorrefmark{1}, 
	LiFang Yang\IEEEauthorrefmark{1},
	Xianglin Huang\IEEEauthorrefmark{1}, 
	Gang Cao\IEEEauthorrefmark{1},
	Tao Zhulin\IEEEauthorrefmark{1},
	Zhengyang Du\IEEEauthorrefmark{2},
	Jing An\IEEEauthorrefmark{1}
	
	\thanks{Xianglin Huang is the corresponding author(e-mail:huangxl@cuc.edu.cn).}
	
	\IEEEauthorblockA{\IEEEauthorrefmark{1}State Key Laboratory of Media Convergence and Communication,Communication University of China}
	
	\IEEEauthorblockA{\IEEEauthorrefmark{2}Columbia University in the City of New York}
}

\maketitle

\begin{abstract}

At present, attention mechanism has been widely applied to the fields of deep learning models. Structural models that based on attention mechanism can not only record the relationships between features’ position, but also can measure the importance of different features based on their weights. 
By establishing dynamically weighted parameters for choosing relevant and irrelevant features, the key information can be strengthened, and the irrelevant information can be weakened. 
Therefore, the efficiency of deep learning algorithms can be significantly elevated and improved. 
Although transformers have been performed very well in many fields including reinforcement learning, there are still many problems and applications can be solved and made with transformers within this area. 
MARL (known as Multi-Agent Reinforcement Learning) can be recognized as a set of independent agents trying to adapt and learn through their way to reach the goal. 
In order to emphasize the relationship between each MDP decision in a certain time period, we applied the hierarchical coding method and validated the effectiveness of this method. This paper proposed a hierarchical transformers MADDPG based on RNN which we call it Hierarchical RNNs-Based Transformers MADDPG(HRTMADDPG). 
It consists of a lower level encoder based on RNNs that encodes multiple step sizes in each time sequence, and it also consists of an upper sequence level encoder based on transformer for learning the correlations between multiple sequences so that we can capture the causal relationship between sub-time sequences and make HRTMADDPG more efficient. 

\end{abstract}


\IEEEpeerreviewmaketitle

\section{Introduction}

Information has always played an important role in people's daily communication, such as image processing, information recognition, intelligent computing, automatic control, etc., which are all researched on the basis of information.
However, the cumbersome, large and vague data often hinders experts and scholars in exploring the content of the information, so some science and technology about processing information have sprung up.
Artificial Intelligence(AI), mainly represented by deep learning, began to be active in people's sight.
In recent years, deep learning has always been a leader in the field of AI, and has a wide range of applications in pattern recognition, computer vision, and Natural Language Processing(NLP).
The idea of deep learning comes from the study of artificial neural networks, which are inspired by real brain structures, induced many types of neural networks.
Each neuron in the network can receive, process input signals and send output signals.
The relationship of each neuron to the connections of other neurons is evaluated by a real number called the weight coefficient, which reflects the importance of a given connection in the neural network.
The form of deep learning is represented by the neural network structures and people can learn a large number of information features through the input and output connections between each layer. 

The attention mechanism\cite{bahdanau2016neural, Zhulin} is the core technique widely used in NLP, statistical learning, image processing, speech recognition and computer science after the rapid development of deep learning.
According to the research on human attention, the attention mechanism is proposed, which is essentially to realize the efficient allocation of information processing resources.
Transformer\cite{transformer} abandons the traditional CNN and RNN and the whole network structure is composed entirely of attention mechanism.
Transformer has achieved excellent results in many fields and has solved the problem with a Recurrent Neural Networks(RNNs) which are limited to sequence.
In particular, Transformer Architecture\cite{transformer} has achieved breakthrough success in a number of areas:
Natural language domain\cite{bert},
Machine translation field\cite{tensor2tensor}.
In addition to the natural language fields described above, Transformer networks have achieved a lot of art-of-state in other fields, such as target detection\cite{facebookendtoend}, automated driving\cite{transformer_lane}.
However, Transformer is not widely used in the field of Deep Reinforcement Learning(DRL), in which GTRXL\cite{pmlr-v119-parisotto20a} proposes a stable learning network structure,
and Uddeshya\cite{upadhyay2019transformer} combines the Transformer with DQN and applies it to the Cartpole Environment of OpenAI GYM.

One of the big challenges in the field of Reinforcement Learning (RL) is to develop an efficient swarm intelligence based multi-agent system and to optimize the involving tasks\cite{9043893}. 
Therefore, by combining the excellent achievements of Transformer in many fields and integrating the Multi-agent Deep Reinforcement Learning (MARL) of Transformer Architecture, we can achieve great research significance.

Time sequence analysis is an important work in many fields.
Conventional neural network prediction data often use RNN to forecast, due to the number of training layers and long distance sequences, this method often has gradient explosion and gradient disappearance problems\cite{S0218488598000094}.
Thus a variant of RNN that combines the idea of an attention mechanism appears Long-Short-Term Memory Artificial Neural Network (LSTM).
LSTM was originally proposed by Hochreiter and Schmidhuber in 1997\cite{lstm}. 
Recently, it has been improved and promoted by Alex Graves\cite{graves2014generating}, making it more flexible to be used in a variety of occasions.
LSTM is still essentially a structure of RNN\cite{GravesLstm}, but it can solve the gradient disappearance problem in RNN because it has a uniquely designed "gate" structure (input gate, forgetting gate, and output gate).

The way of making multi-agents cooperate or compete with each other in a team can be regarded as a Partially Observable Markov Decision Process(POMDP) problem.
The POMDP simulation agent decision-making procedure assumes that the system dynamics are determined by the MDP, but the agent cannot directly observe the state.
In the field of single agent, researchers have done a lot of work. 
Deep Q-learning(DQN)\cite{dqn} solves the problem of the dimension disaster of Q-learning. 
The DDPG algorithm proposed by Lillicrap\cite{ddpg} makes RL deal with the problem of continuous action.
With the continuous updating iteration of actor-critic framework\cite{actor_critic} algorithm, Heess\cite{rdpg} combines DDPG with LSTM and proposes RDPG, which optimizes the efficiency of RL to deal with continuous sequences.
In the RL task of single agent, DDPG algorithm performs well in the environment of single agent.
However, in MARL, the multi-agent environment is unstable compared with the single-agent environment, so it brings many difficulties in the process of RL.
Based on the basic framework of DDPG, Ryan\cite{maddpg} applies the Deep Deterministic Policy algorithm to multi-agent environment, and proposes the Multi-Agent Deep Deterministic Policy Gradient(MADDPG) algorithm.
Similarly, Wang\cite{wang2020rmaddpg} combines LSTM with MADDPG and makes RMADDPG work well in the POMDP environment, while Shariq Iqbal\cite{attention_maddpg} associates attention with MADDPG and works well in the Cooperative Treasure Collection and roter-tower self-built environments.
The rest, such as\cite{Recurrent-MADDPG, jiang2018learning, wei2020rmaddpg}, 
have made great improvements in MADDPG and improved the performance of multi-agents in different fields.

We found that transformer has many problems in the process of combining with RL, especially position encoding.
In NLP, position encoding is encoded according to the number of words, and RL does not have a corresponding word library.
The position encoding problem also exists in other supervised learning fields, 
for example, Xia\cite{xia2019rthn} proposes a Hierarchical Network Structure (RTHN) based on RNN and Transformer to model and classify the relationship between multiple clauses in a document in a joint framework,
Liu\cite{liu2019hierarchical} proposes a Transformer architecture that encodes documents in a hierarchical manner 
and Pappagari\cite{9003958} encodes documents in a hierarchical manner.
Inspired by Hierarchical Transformer Network, in this paper we combine Transformer Architecture\cite{transformer} with Multi-agent Deep Deterministic Policy Ggradient (MADDPG)\cite{maddpg}.
The Transformer Encoding schema is optimized according to the hierarchical network structure of Xia\cite{xia2019rthn} and Liu\cite{liu2019hierarchical}.

In this work, we propose a multiagent hierarchical network architecture based on RNN and Transformer, named Hierarchical RNNs-Based Transformers MADDPG(HRTMADDPG), to model the relations between multiple time sequences in a multiagent environment and classify them synchronously in a joint framework. 
HRTMADDPG is composed of two layers: 
1) The lower layer is a step-level encoder consisting of multiple RNNs, each of which corresponds to one time sequence, in turn encodes the steps in the time sequences and combines them to obtain the time sequences representation; 
2)The upper layer is a time sequence-level encoder based on a stacked Transformer, where the time sequences representations are repeatedly learned and updated by incorporating the relations between multiple sequences, and finally feed to a softmax layer for synchronous classification.

We further propose ways to encode the relative position and gain further improvements.
On one hand, the attention mechanism in Transformer learns the correlation between sequences. On the other hand, the encoding of hierarchical prediction predicts the relationship between sequences.

The main contributions of this paper are as follows:
\begin{enumerate}
	\item We propose a new hierarchical network architecture based on RNNs and Transformer for the multiagent task. To the best of our knowledge, it is the first time that Transformer has been used to solve multiagent problems. It demonstrates excellent performance in learning the correlation between multiagents.
	\item  We further encode the relative position and global predication information into the Transformer framework. It can capture the causality between sequence and achieve extra improvements.
	\item The effectiveness of our model is demonstrated on the benchmark MADDPG. We finally achieve the best performance.
\end{enumerate}

The remainder of this manuscript is structured as follows. 
Section II introduces some background knowledge of MDPs and POMDPs, MADDPG, recurrent MADDPG.  
The proposed algorithm is demonstrated in Section III. 
Simulation results and discussions are presented in Section IV. 
Section V concludes this paper and envisages some future work.

\section{Background}
\subsection{POMDP}

The POMDP model is used to describe the partially observable Markov process with hidden system states and uncertain behavior effects.
The POMDP model uses a six-tuple $S, A, T, \Omega, R, O$.

\begin{enumerate}
	\item $S$: Represents the state space, in which each state contains all the information of the environment. The form and scope of the state space are unknown to the agent.
	\item $A$: Represents the action space of the agent, including all feasible actions of the agent;
	\item $T$: $S*A \to \triangle S$ represents the state transfer function of the environment. 	
	Since the state contains all the information of the environment, the transfer equation has Markov property, that is, to calculate the state at the next moment, only the state at the current moment is needed, which has nothing to do with the previous states.
	\item $\Omega$: Represents the observation function, it maps environment state to the observation space of the agent;
	\item $R$: Represents the reward function, which measures the degree of completion of the decision-making task;
	\item $O$: Represents the observation space and is the source of agent information. 
	Agents can only indirectly infer the true state of the environment through observation.
\end{enumerate}

In POMDP, the task of the agents is to learn the policy function $\pi(a_t|o_0...o_t)$ by continuously interacting with the environment,
so that the trajectory can be obtained by following the function $\tau=\{o_0,a_0,r_1,o_1,a_1,r_2...\}\in \mathcal{T}$, then the largest expected value of cumulative reward can be calculated by the following function:

\begin{equation}
	\pi= \mathbb{E}\Big[\sum_{t=1}^{\infty}\gamma^{t}r_t \Big]
\end{equation}

\subsection{Multiagent Reinforcement Learning}

At present, the main algorithms of RL can be roughly divided into two categories: one is value-based algorithm (Value-Based), and the other is policy-based algorithm (Policy-Based). 
At the same time, it can also be classified by model, and divided into model-based algorithms and model-independent methods.
RL algorithms are updated quickly, and excellent algorithms such as Deep Q-learning\cite{dqn}, Sarsa, Policy Gradients, etc. have appeared.
In the field of games, such as Go game\cite{AlphaGo}, "StarCraft"\cite{AlphaZero}, etc. have shown considerable results.
At the same time, these algorithms have a wide range of potential applications in fields such as autonomous vehicles\cite{kiran2021deep}, drone navigation\cite{2019icrae}, trajectory planning\cite{wei2020rmaddpg}, etc.
In addition to the importance of RL in the field of control, it has also been widely used in computer vision\cite{2309945}, natural language processing\cite{madureira2020overview}, audio signal processing\cite{latif2021survey}, recommendation system\cite{afsar2021reinforcement} and other fields .

The depth of deep learning algorithms Deep q-learning(DQN)\cite{dqn} solves the problem of Q-learning "dimension disaster", but it cannot directly solve the problem of continuous actions.
On the basis of the work of the DQN algorithm, the Double DQN algorithm proposed by Hasselt et al.\cite{Silver2016} combines the idea of DQN, and gives a general explanation of the overestimation and mathematical proof of the solution, and finally has a super high score in the Atari game Experimental performance.
In order to make the deterministic policy gradient have a satisfactory exploration effect,
Silver\cite{pmlr-v32-silver14} uses the off-policy learning algorithm, which selects actions based on random behavior policy. 
Silver uses a deterministic policy gradient to learn an actor-critic algorithm for estimating action-value.
Furthermore, Mnih\cite{a3c} proposed an asynchronous RL solution and applied it to many classic RL algorithms, so that the training speed and learning effect are significantly improved.
On this basis, the participants of the Deepmind team proposed Deep deterministic policy gradient (DDPG)\cite{ddpg}, which is a RL algorithm that solves the problem in the continuous action space.
DDPG is a model-free DRL algorithm for policy learning participants,
which uses a deep deterministic neural network model to achieve learning problems in the continuous weighted action space.
However, the Policy Gradient algorithm is very sensitive to the step size, but it is difficult to choose a suitable step size. 
it is not conducive to learning when the difference between the previous and the current policy changes during the training process is too large. 
The PPO\cite{ppo} and TRPO\cite{trpo} proposed by Schulman solve the problem that the step length is difficult to determine in the Policy Gradient algorithm.
Based on the Actor-Critic(AC) method, OpenAI proposed a MARL algorithm MADDPG\cite{maddpg},
MADDPG considers the relationship between various agents, and uses centralized training and decentralized execution ideas to achieve significant results.

\subsection{Recurrent MADDPG}

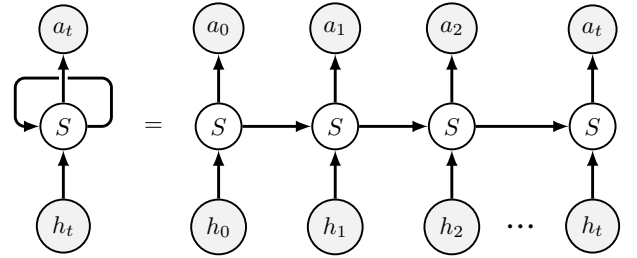
\begin{figure}[!hbtp]
	\centering
	\begin{adjustbox}{width=\textwidth/9*4}%
		\begin{tikzpicture}[item/.style={circle,draw,thick,align=center},
		itemc/.style={item,on chain,join}]
		\begin{scope}[start chain=going right,nodes=itemc,every
		join/.style={-latex,very thick},local bounding box=chain]
		
		\path node (A0) {$S$} node (A1) {$S$} node (A2) {$S$} node[xshift=1em] (At) {$S$};
		\end{scope}
		
		\node[left=1em of chain,scale=1] (eq) {$=$};
		\node[left=2em of eq,item] (AL) {$S$};
		\path (AL.west) ++ (-1em,2em) coordinate (aux);
		\draw[very thick,-latex,rounded corners] (AL.east) -| ++ (1em,2em) -- (aux) |- (AL.west);
		\foreach \X in {0,1,2,t} 
		{
			\draw[very thick,-latex] (A\X.north) -- ++ (0,2em)	node[above,item,fill=gray!10] (h\X) {$a_\X$};
			\draw[very thick,latex-] (A\X.south) -- ++ (0,-2em) node[below,item,fill=gray!10] (x\X) {$h_\X$};
		}
		\draw[white,line width=0.8ex] (AL.north) -- ++ (0,1.9em);
		\draw[very thick,-latex] (AL.north) -- ++ (0,2em) node[above,item,fill=gray!10] {$a_t$};
		\draw[very thick,latex-] (AL.south) -- ++ (0,-2em) node[below,item,fill=gray!10] {$h_t$};
		\path (x2) -- (xt) node[midway,scale=1,font=\bfseries] {\dots};
		\end{tikzpicture}
	\end{adjustbox} 
	\caption{
		Recurrent Multiagent Deep Deterministic Policy Gradient (RMADDPG) algorithm. The observation history is $h$, and $t$ represents the current time step, $s$ represents the current memory state, and $a$ represents the action output.}
	\label{fig:M1}
\end{figure}
In order to address partial observability, we extend MADDPG using recurrent neural networks trained with backpropogation through time.
In order to solve POMDP, the network learns to preserve the limited information about the past, using RNN. 
RNN captures the history of k observations and actions from current time step $t$ to time step $t-k$ (where k is a hyper-parameter specified by the user and is domain dependent).
The architecture of a simple RNN used for this research is described in Fig. 1, where st is the current hidden state, and ht is the current observation. Mathematically, the hidden state is calculated as $s_t=f(Uh_t+Ws_{t-1})$. 
Each observation and the current "memory" are passed through feed-forward neural networks U and W respectively; they are then summed and passed through a non-linear function to create the final output action. This action is then used to take a step in the environment.

The architecture of our proposed algorithm HRTMADDPG, is shown in Figure \ref{HRTMADDPG}, where the actors take
actions based on the output generated by RNN, based on information of past states and observations.
Once we have the state, action, next-state, and reward based on the current policy, the actor and critic values are updated as following:

Update critic by minimizing the loss: 
\begin{equation}
\mathcal{L}(\theta_i)=\frac{i}{S}{{\left( \sum\nolimits_{j}{{{y}^{i}}-Q_{i}^{\mu }(x^j,h^j,a_{1}^{j},...,a_{N}^{j})} \right)}^{2}}
\label{eq_critic_loss}
\end{equation}

\begin{equation}
\begin{split}
{{\nabla }_{{\theta_i}}}J\approx\\
&\frac{i}{S}\sum\limits_{j}{{{\nabla }_{{{\theta }_{i}}}}}{{\mu }_{i}}(s_{i}^{j},h_{i}^{j})\\
&{{\nabla }_{{{a}_{i}}}}Q_{i}^{\mu }(x^j,h^j,a_{1}^{j},...,a_{N}^{j}){{|}_{{{a}_{k}}={{\mu }_{k}}(s_{k}^{j},h_{k}^{j})}}
\label{eq_actor_loss}
\end{split}
\end{equation}

\subsection{Transformer}

\begin{figure}[!t]
	\centerline{\includegraphics[width=3in]{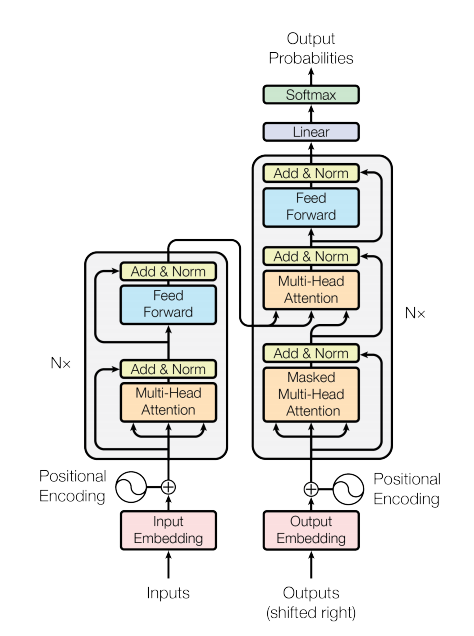}}
	\caption{The Transformer-model architecture, 
		 from Transformer\cite{transformer}.}
	\label{transformer}
\end{figure}

The attention mechanism has been proposed as early as the 1990s, and the Google DeepMind team combined the attention mechanism with RNN for image classification and achieved remarkable results\cite{mnih2014recurrent}. 
In addition, Bahdanau et al.\cite{bahdanau2016neural} used the attention mechanism in NLP, which greatly improved the translation accuracy, and also allowed the attention mechanism to be continuously developed and applied to various fields. 
Analyzing and comparing various application fields of attention mechanism, people have been researching for the purpose of improving efficiency and overcoming the limitations of algorithms such as CNN and RNN, and trying to propose new algorithm structures.
Vaswani\cite{transformer} proposed a Transformer model with self-attention as the basic unit, so that the attention mechanism can be successfully applied.
In the field of multi-agent reinforcement learning, researchers have tried to add an attention mechanism, such as Iqbal\cite{attention_maddpg}, which combines attention with MADDPG and achieved remarkable results.

With the continuous development and research of Attention, the Google DeepMind recently replaced the Seq2Seq problem with transformer, and replaced LSTM with self-attention, and achieved unprecedented results in tasks such as translation\cite{bert}.
The Transformer system based on the self-attention mechanism includes an encoder and a decoder. 
Unlike machine translation based on RNN, the Transformer encoder and decoder are composed of a novel attention mechanism and a feedforward neural network.
Transformer named this novel attention mechanism(Multi-head Attention, MHA), the overall architecture of Transformer is shown in Figure \ref{transformer}.

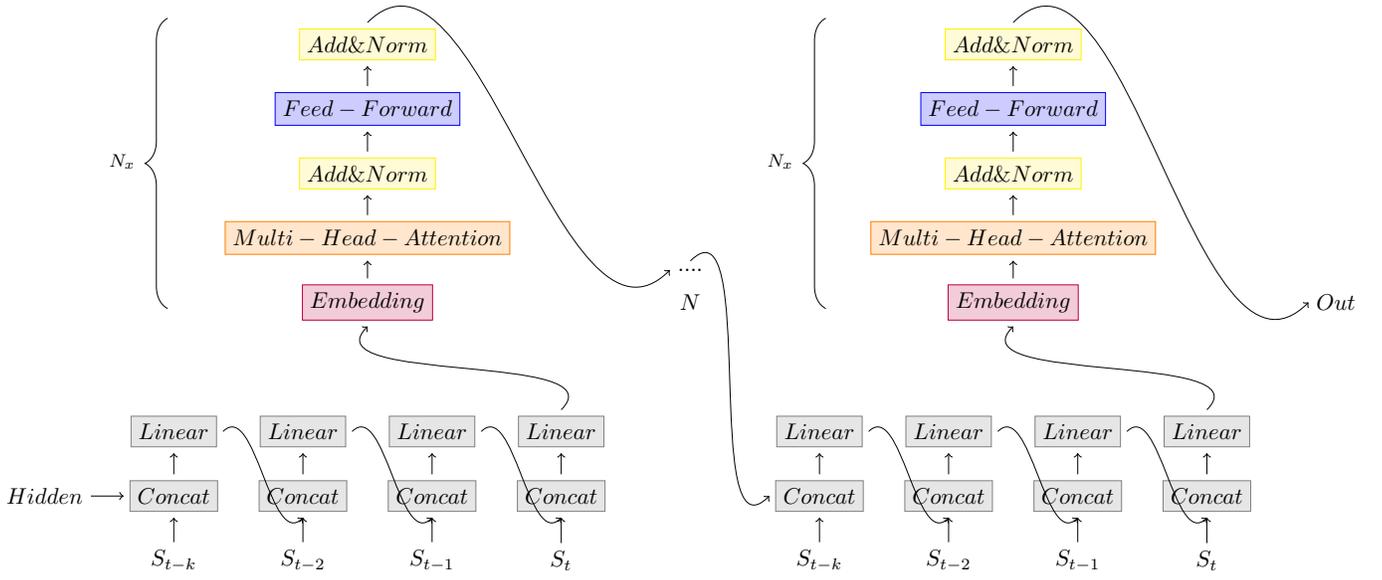
\begin {figure*}[!hbtp]
\centering
\begin{adjustbox}{width=\textwidth}
	\begin{tikzpicture}[
		hid/.style 2 args={
			rectangle split,
			rectangle split horizontal,
			draw=#2,
			rectangle split parts=#1,
			fill=#2!20,
			outer sep=1mm}]
		
		\foreach \i [count=\step from 1] in {$S_{t-k}$,$S_{t-2}$,$S_{t-1}$,$S_t$}
		\node (i\step) at (2*\step, -2) {\i};
		
		\node[align=center] (o0) at (0, -1) {$Hidden$};
		\foreach \step in {1,...,3} {
			\node[hid={1}{gray}] (h\step) at (2*\step, 0) {$Linear$};
			\node[hid={1}{gray}] (e\step) at (2*\step, -1) {$Concat$};    
			\draw[->] (i\step.north) -> (e\step.south);
			\draw[->] (e\step.north) -> (h\step.south);
		}
		
		
		\node[hid={1}{gray}] (h4) at (2*4, 0) {$Linear$};
		\node[hid={1}{gray}] (e4) at (2*4, -1) {$Concat$};   
		
		\node[hid={1}{yellow}] (s44) at (2*2.5, 6) {$Add \& Norm$};
		\node[hid={1}{blue}] (s43) at (2*2.5, 5) {$Feed - Forward$};
		\node[hid={1}{yellow}] (s42) at (2*2.5, 4) {$Add \& Norm$};
		\node[hid={1}{orange}] (s41) at (2*2.5, 3) {$Multi-Head-Attention$}; 
		\node[hid={1}{purple}] (s4) at (2*2.5, 2) {$Embedding$}; 
		
		\draw[->] (s41.north) -> (s42.south);
		\draw[->] (s42.north) -> (s43.south);
		\draw[->] (s43.north) -> (s44.south);
		
		\path (h4.north) edge[->,out=45,in=225] (s4.south);

		\draw[->] (s4.north) -> (s41.south);
		\draw[->] (e4.north) -> (h4.south);
		
		\draw[->] (i4.north) -> (e4.south);

		\node[align=center] (o40) at (2*5, 2.5) {$....$};
		\node[align=center] (o41) at (2*5, 2) {$N$};
		
		\foreach \i [count=\step from 5] in {$S_{t-k}$,$S_{t-2}$,$S_{t-1}$,$S_t$}
		\node (i\step) at (2*\step + 2, -2) {\i};
		
		\node[hid={1}{yellow}] (s84) at (2*6.5 + 2, 6) {$Add \& Norm$};
		\node[hid={1}{blue}] (s83) at (2*6.5 + 2, 5) {$Feed - Forward$};
		\node[hid={1}{yellow}] (s82) at (2*6.5 + 2, 4) {$Add \& Norm$};
		\node[hid={1}{orange}] (s81) at (2*6.5 + 2, 3) {$Multi-Head-Attention$}; 
		\node[hid={1}{purple}] (s8) at (2*6.5 + 2, 2) {$Embedding$}; 
		
		\draw[->] (s81.north) -> (s82.south);
		\draw[->] (s82.north) -> (s83.south);
		\draw[->] (s83.north) -> (s84.south);

		\foreach \step in {5,...,7} {
			\node[hid={1}{gray}] (h\step) at (2*\step + 2, 0) {$Linear$};
			\node[hid={1}{gray}] (e\step) at (2*\step + 2, -1) {$Concat$};    
			\draw[->] (i\step.north) -> (e\step.south);
			\draw[->] (e\step.north) -> (h\step.south);
		}
		
		
		\node[hid={1}{gray}] (h8) at (2*8 + 2, 0) {$Linear$};
		\node[hid={1}{gray}] (e8) at (2*8 + 2, -1) {$Concat$};   
		
		\draw[->] (s8.north) -> (s81.south);
		\draw[->] (e8.north) -> (h8.south);
		\path (h8.north) edge[->,out=45,in=225] (s8.south);
		
		\draw[->] (i8.north) -> (e8.south);
		
		\path (s44.north) edge[->,out=45,in=225] (o40.west);
		\path (o40.north) edge[->,out=45,in=225] (e5.west);
		
		\foreach \step in {1,...,3} {
			\pgfmathtruncatemacro{\next}{add(\step,1)}
			\path (h\step.east) edge[->,out=45,in=225] (e\next.south);
		}
		
		\foreach \step in {5,...,7} {
			\pgfmathtruncatemacro{\next}{add(\step,1)}
			\path (h\step.east) edge[->,out=45,in=225] (e\next.south);
		}
		
		\node[align=center] (h9) at (2*9 + 2, 2) {$Out$};
		\path (s84.north) edge[->,out=45,in=225] (h9.west);
		\draw[->] (o0.east) -> (e1.west);
		
		\draw [decorate,decoration={brace,amplitude=10pt},xshift=40pt,yshift=40pt]
		(0.5,0.5) -- (0.5,5.0) node [black,midway,xshift=-20pt,yshift=1pt]
		{\footnotesize $N_x$};
		
		\draw [decorate,decoration={brace,amplitude=10pt},xshift=330pt,yshift=40pt]
		(0.5,0.5) -- (0.5,5.0) node [black,midway,xshift=-20pt,yshift=1pt]
		{\footnotesize $N_x$};
		
	\end{tikzpicture}
\end{adjustbox}

\caption{The framework of HRTMADDPG model.}
\label{HRTMADDPG}

\end{figure*}

\begin{algorithm}
\caption{Tansformers Multiagents Deep Deterministic Policy Gradient Based on RNN for N Agents}
\begin{algorithmic}[1]
	\For{episode = 1 to M}
	\State Initialize a random process $\mathcal{N}$ for action exploration
	\State Receive initial state $s$
	initialize empty history $h$
	\For{t = 1 to max-episode-length}
	\State for each agent i select action:
	\State $a_{i}= \mu_{\theta_i}(o_i, h_i) + \mathcal{N}_{i}$
	\State Execute actions $a = (a_{i},..., a_{N})$
	\State Get reward $r$
	\State Get new state ${s}'$
	\State Get new history ${h}'$
	\State store $(s, h, a, r, s', h')$ in replay buffer $\mathcal{D}$
	\State $s \gets s'$ 
	\State $h \gets h'$ 
	\For{agent $i = 1$ to N}
	\State Sample a random mini-batch of $\mathcal{S}$ samples
	\State $(s^j, a^j, r^j, s'^j, h'^j)$ from $\mathcal{D}$
	
	\State $x = s^j,h^j,a_{1}^{j},...,a_{N}^{j}$
	\State $x' = {s^j}',{h^j}', {a_1}',...,{a_N}'$
	\For{t = 1 to hierarchical-layers}
	\State $x' = Attention(L(H(x')))$
	\EndFor
	
	\State $\mathcal{Y}^i=r_i^j +\gamma Q_i^{{\mu}'}(x'){{|}_{{{a}_{k}}^{\prime }={{\mu }_{k}}^{\prime }(s_{k}^{j},h_{k}^{j})}}$
	\State Update critic by minimizing the loss: 
	\State $\mathcal{L}(\theta_i)=\frac{i}{S}{{\left( \sum\nolimits_{j}{{{y}^{i}}-Q_{i}^{\mu }(x)} \right)}^{2}}$
	\State Update actor:
	\State ${{\nabla }_{{\theta_i}}}J\approx$ 
	\State $\qquad\qquad\frac{i}{S}\sum\limits_{j}{{{\nabla }_{{{\theta }_{i}}}}}{{\mu }_{i}}(s_{i}^{j},h_{i}^{j})$
	\State \qquad\qquad${{\nabla }_{{{a}_{i}}}}Q_{i}^{\mu }(x){{|}_{{{a}_{k}}={{\mu }_{k}}(s_{k}^{j},h_{k}^{j})}}$
	\EndFor
	\State Update target network parameters 
	\State for each agent $i$
	\State $\theta_{i}' \gets \tau\theta_{i} + (1-\tau)\theta_{i}'$
	\EndFor
	\EndFor
\end{algorithmic}
\end{algorithm}

\section{Methods}

In principle, the RNN model does not require position coding, and it has the possibility of learning position information in its structure.
Therefore, if you connect a layer of RNN after the input, and then connect to the Transformer, then theoretically there is no need to add position codes.
In the same way, we can also use the RNN model to learn an absolute position coding, such as starting from a vector, and obtaining the coding vector of each position through a recursive format.
Liu\cite{pmlr_v119_liu20n} proposed the use of Neural Ordinary Differential Equations(ODE) to model position codes. This scheme is called FLOATER.
FLOATER is a recursive model, and functions can be modeled by neural networks. Therefore, this type of differential equation is also called a neural differential equation, and work on it has gradually increased recently.
Theoretically speaking, position coding based on recursive models also has better extrapolation, and it is also more flexible than trigonometric position coding. Liu\cite{pmlr_v119_liu20n} proved the position coding of trigonometric function It is a special solution of FLOATER. However, it is obvious that recursive position coding sacrifices a certain degree of parallelism and may bring a speed bottleneck.

\subsection{Transformer Position Embedding}

In Tranformer\cite{transformer}, Vaswani uses sine and cosine to encode the position of words, which is different from NLP.
In a multi-agent environment, the input comes from the state of each agent. Unlike NLP, MARL does not have a fixed word library.
Vaswani\cite{transformer} and use sine and cosine to calculate positional embeddings.

\begin{equation}
	\begin{split}
		e_p[i]=sin(p/10000^{2i/d}) \\
		e_p[2i + 1]=cos(p/10000^{2i/d})
	\end{split}
	\label{eq1}
\end{equation}

\subsection{RNNs Position Embedding}

As has mentioned in the Introduction, 
MARL can be regarded as the optimization problem of a group of independent agents in a limited time. 
In order to emphasize the relationship between each MDP decision in a certain time period, we adopted the hierarchical coding method and verified the effectiveness of this method.
This framework normally has only one encoding layer (step-level encoder).
In this work, the proposed HRTMADDPG model is a 2-layer hierarchical network containing not only step-level encoders at the lower layer but also a sequence level encoder at the upper layer.

Each time series corresponds to an RNN module, which is the information accumulated at each step of the sequence.
The low level contains $N-th$ RNNs encoders, $N-th$ RNNs correspond to the state of $N-th$ time point ending at the current time, where $i-th$ represents the distance from the current time point Is the state of the time point of $i$.
Based on RNN to obtain $i-th$, the corresponding hidden state $h_i$ can be obtained, and then the step-level attention mechanism is adopted to obtain the sequence representation $r_i$ through the incremental sum of the hidden state of all steps in the sequence.
Here, we define$\{p_i\}_{i \in \{1,2,...N\}}$ as a position code within a time sequence.

\begin{equation} 
	p_{i+1} = RNN(h_i,p_i)
	\label{rnns_encoder}
\end{equation}

\subsection{Transformer encoder}

In the Transformer \cite{transformer} encoder, the Transformer is composed of $ N$ layers, each of which has a multi-head self-attention and fully connected feedforward network.
In our work, as shown in the formula \ref{rnns_encoder}, the shallow steps are encoded by RNNs, and the subsequent high-level time series are encoded by Transformer.
Figure \ref{HRTMADDPG} provides a schematic diagram of attention between sequences.

\begin{equation}
	H=\text{LayerNorm}\left(\mathcal {D}^{(k)}+\mathrm{MultiHead}\left(\mathcal {D}^{(k)}\right)\right)
\end{equation}

\begin{equation} 
	{L}^{(k)}=\text{LayerNorm}(H+\mathrm{MLP}(H))
\end{equation}

is calculated as follows:

Among them, LayerNorm is the layer normalization of dating in \cite {Layer_Normalization}, MLP represents a two-layer feedforward network with ReLU activation function, and MultiHead represents the multi-head attention mechanism proposed in \cite{transformer}.
Multi-head attention applied to MARL $\lbrace \mathcal {D}^{k}\rbrace$
The calculation is as follows:

\begin{equation} 
	\mathrm{MultiHead}= \text{Concat} \left(\mathrm{head}_{1}, \ldots , \mathrm{head}_{\mathrm{h}}\right) W^{O} 
\end{equation}

\begin{equation} 
	\begin{aligned} 
		\text{ where }&Q = \mathcal {D}^{(k)} W_{i}^{Q},\\ &K = \mathcal {D}^{(k)} W_{i}^{K},\\ &V = \mathcal {D}^{(k)} W_{i}^{V} 
	\end{aligned} 
\end{equation}

where
$W_{i}^{Q},W_{i}^{K},W_{i}^{V} \in \mathbb{R}^{d_s \times d_s}$
are weight metrics, and the attention is computed as

\begin{equation} 
	\text{ Attention }({Q},{K},{V})=\operatorname{softmax}\left(\frac{{QK}^{\top }}{\sqrt {d_{s}}}\right)\mathbf {V} 
\end{equation}

for some input query, key and value matrices 
$Q,K,V \in \mathbb{R}^{M \times d_s}$. 
The $h$ outputs from the attention calculations are concatenated and transformed using a output weight matrix 
$ W^{O} \in \mathbb{R}^{d_sh \times d_s}$.

\subsection{Hierarchical Transformers}

The location encoding of Transformer has been used to represent location information, but it reflects the absolute location information of steps in a sequence and is not applicable to state changes in MDP.
In our task, relative position is more important than absolute position, because changes in the environment are more strongly influenced by changes in nearby states.Therefore, the relative position time sequences can more accurately express the state of the current environment.
In this work, we use Hierarchical Transformers$\{T_1, T_2, ..., T_N\}$ to encode such relative position information, $T_i$ is each layer of Transformer encoder, and $HT_i$ is the last The calculation result:

\begin{equation} 
	HT(x) = T_N \circ T_{N-1} \circ ... \circ T_1(x)
	\label{hierarchical_embedding}
\end{equation}

\section{Experiments}

\begin{figure}[!h]
	\centerline{\includegraphics[width=3in]{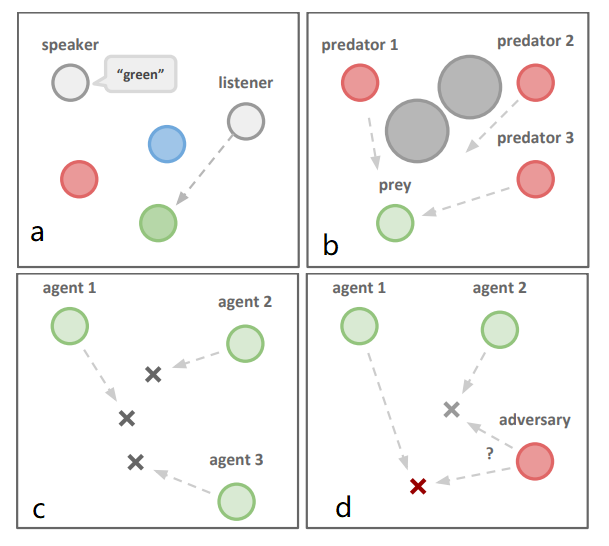}}
	\caption{Illustrations of the experimental environment and some tasks we consider, including a) Cooperative Communication b) Predator-Prey c) Cooperative Navigation d) Physical Deception\cite{maddpg}.}
	\label{mape}
\end{figure}

In order to verify the effectiveness of our algorithm, we use the Multi-Agent Particle Environment(MAPE)\cite{mape} game Environment proposed by OpenAI. MAPE provides two two-dimensional scenes of continuous and discrete modes, and agents can cooperate with each other or have opposing behaviors. At the same time,MAPE also considers the mutual communication between agents.
As shown in the Figure \ref{mape}.

This paper conducts experiments on MADDPG, RMADDPG, and HRTMADDPG of different levels in four different test environments. 25,000 rounds of training and 5,000 rounds of testing are used, and each round has 25 steps in length. 

\subsection{Mixed Cooperative-Competitive Environments}

\subsubsection{Cooperative navigation}
	
Cooperative navigation.This environment contains three Landmarks for three agents, each of which needs to navigate cooperatively and capture each Landmark one by one, while agents need to avoid colliding with each other during this process, the team reward function is,

\begin{equation} 
R = \sum_{t=1}^{}d(p_i,g)
\label{distance_reward}
\end{equation}

\begin{figure}[!h]
	\centerline{\includegraphics[width=3.5in]{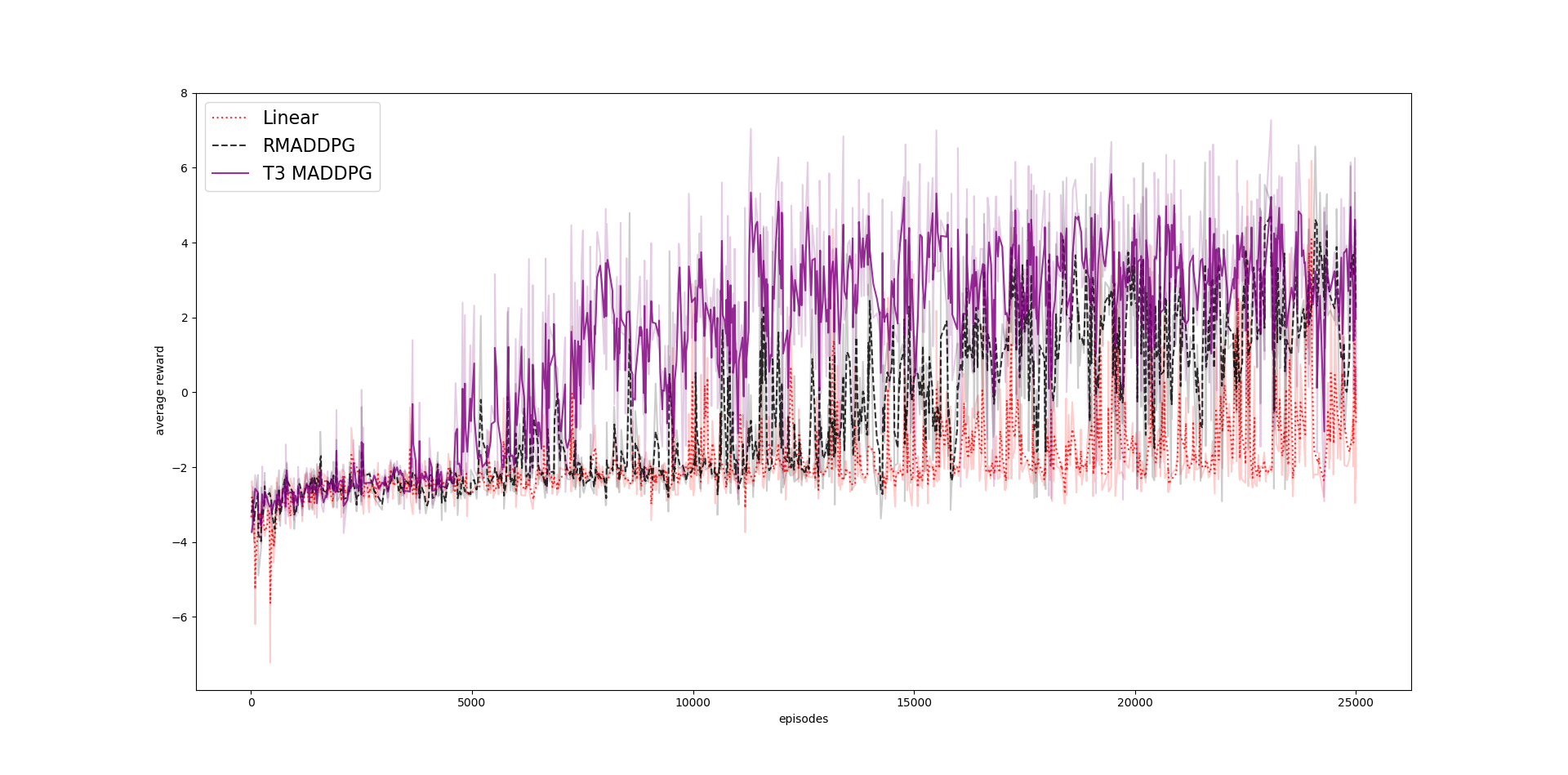}}
	\caption{Cooperative Navigation}
	\label{cooperative_navigation}
\end{figure}

Each agent needs to navigate from position $p_i$ to the goal position $g=(g_x, g_y)$, where $d$ is Euclidean distance.
As shown in Figure \ref{cooperative_navigation} and Table \ref{cooperative_navigation_table}, the Cooperative Navigation experimental environment, the speed of reward curve fitting is greatly improved with the deepening of Transformer Encoding level.
The average HRTMADDPG reward has been increased from -2 to +2 for MADDPG, and the average test reward has been increased.

\begin{table}[!h]
	\centering
	\caption{Reward of the proposed DRL framework at cooperative navigation for each agent}
	\label{cooperative_navigation_table}
	\begin{tabular}{@{}cccc@{}}
		& \multicolumn{3}{c}{Cooperative Navigation} \\ \cmidrule(l){2-4} 
		& agent1  				& agent2     			& agent3    		\\ \cmidrule(l){2-4} 
		MADDPG    &  -0.7750366				& -0.7598686  			& -0.7728286		\\ \cmidrule(l){1-4} 
		RMADDPG  &  3.318712				& 3.316456				& 3.318392			\\ \cmidrule(l){1-4} 
		T1-HRTMADDPG &  3.231071				& 3.228775 				& 3.231207			\\ \cmidrule(l){1-4} 
		T2-HRTMADDPG &  3.326575				& 3.331479 				& 3.328119			\\ \cmidrule(l){1-4} 
		T3-HRTMADDPG &  3.351472				& 3.353712 				& 3.351648			\\ \cmidrule(l){1-4} 
		T4-HRTMADDPG &  3.322668				& 3.32982				& 3.321108			\\ \cmidrule(l){1-4} 
		T5-HRTMADDPG &  \textbf{3.355197}		& \textbf{3.354853}  	& \textbf{3.352453}	\\ \cmidrule(l){1-4}
	\end{tabular}
\end{table}
	
\subsubsection{Physical deception}
	
This environment contains four agents, three of which are positive agents and one is negative agents. Just like Cooperative navigation, there are also three Landmarks in this environment. The difference is that the opponent will try to navigate to one of the Landmarks if the positive agent does not know the intention of the landmark location that the negative agent is trying to reach.
Thus, this environment involves both cooperation and competition between agents.
The reward function for the negative agent is same to  \ref{distance_reward}, on the other hand, the reward function for the positive agent is the difference between two team.

\begin{equation} 
R = \sum_{t=1}^{}d(positive_i,g) - \sum_{t=1}^{}d(negative_i,g)
\end{equation}

\begin{figure}[!h]
	\centerline{\includegraphics[width=4in]{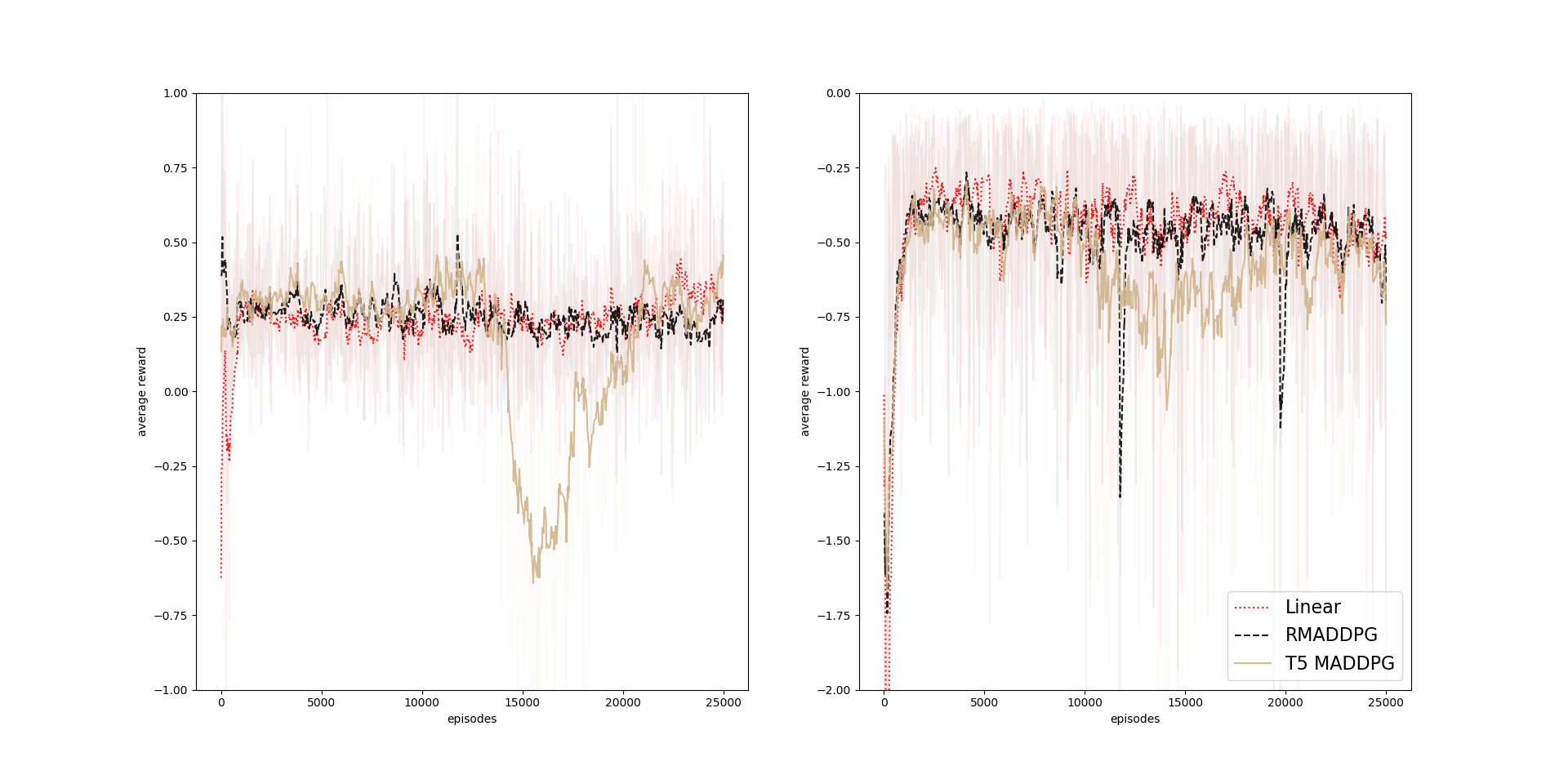}}
	\caption{Physical Deception}
	\label{physical_deception}
\end{figure}

As shown in Figure \ref{physical_deception} and table \ref{physical_deception_table} can be seen,
For the sake of statistics, we took the average reward of the two agents on the positive side and only one agent on the negative side.
Under the experimental environment of Physical deception, with the deepening of the level of Transformer Encoding,
The average reward value of HRTMADDPG training and the average reward value of test are both increasing and gradually approaching the performance of RMADDPG.
However, both RMADDPG and HRTMADDPG fluctuated greatly in the average reward value under this environment.

\begin{table}[!h]
	\centering
	\caption{Reward of the proposed DRL framework at physical deception for each agent}
	\label{physical_deception_table}
	\begin{tabular}{@{}cccc@{}}
		& \multicolumn{3}{c}{Physical Deception} \\ \cmidrule(l){2-4} 
		& agent1  				& 	agent2     			& agent3    				\\ \cmidrule(l){2-4} 
		MADDPG    &  \textbf{-0.441903484} 	& 0.243341461			& 0.243341461			\\ \cmidrule(l){1-4} 
		RMADDPG  &  -0.474591971 			& \textbf{0.25247941}	& 0.25247941			\\ \cmidrule(l){1-4} 
		T1-HRTMADDPG &  -0.632189044 			& 0.220001727			& 0.220001727			\\ \cmidrule(l){1-4} 
		T2-HRTMADDPG &  -0.591533075 			& 0.229312382			& 0.229312382			\\ \cmidrule(l){1-4} 
		T3-HRTMADDPG &  -0.636275915 			& 0.235282304			& 0.235282304			\\ \cmidrule(l){1-4} 
		T4-HRTMADDPG &  -0.532002985 			& 0.262908456			& \textbf{0.262908456}	\\ \cmidrule(l){1-4} 
		T5-HRTMADDPG &  -0.565551017 			& 0.186596965			& 0.186596965 			\\ \cmidrule(l){1-4} 
	\end{tabular}
\end{table}
	
\subsubsection{Cooperative communication}
	
The environment consists of two agents named Speaker and Listener. The Listener needs to navigate to a landmark of a specific color under the guidance of the Speaker.
The speaker determines the color of the output Landmark based on the listener's behavior.
the reward function for the environment of Cooperative communication is depend on the Euclidean distance between the goal position and listener.

\begin{figure}[!h]
	\centerline{\includegraphics[width=4in]{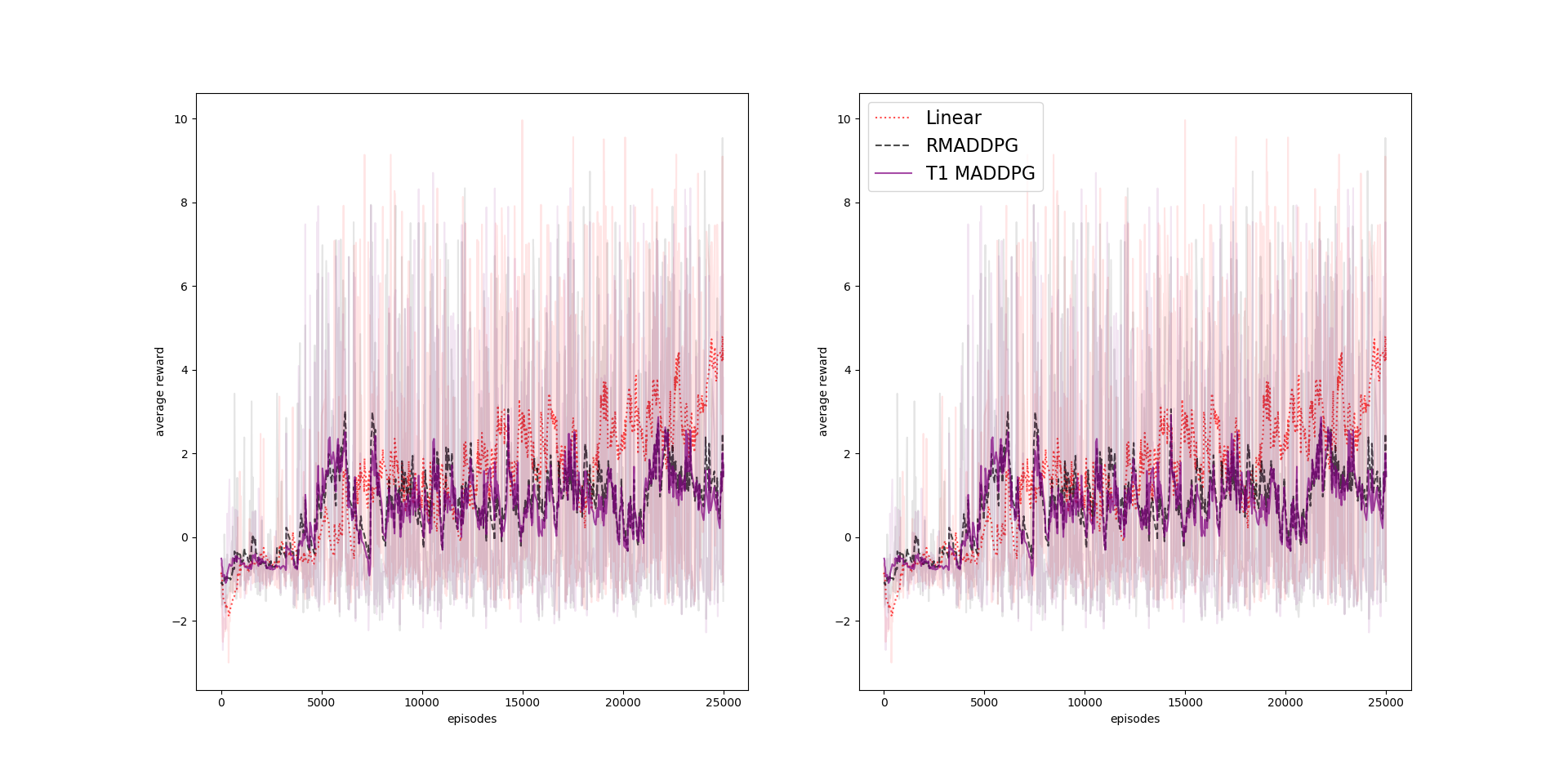}}
	\caption{Cooperative communication}
	\label{cooperative_communication}
\end{figure}

As shown in Figure \ref{cooperative_communication} and table \ref{cooperative_communication_table}, the experimental environment of cooperative communication was accomplished with the deepening of Transformer Encoding level.
The average reward value of HRTMADDPG training and the average reward value of test are both increasing and gradually approaching the performance of RMADDPG.
However, both RMADDPG and HRTMADDPG fluctuated greatly in the average reward value under this environment.
In the test environment, the average reward value of positive and negative agents is larger under MADDPG algorithm.

\begin{table}[!h]
	\centering
	\caption{Reward of the proposed DRL framework at cooperative communication for each agent}
	\label{cooperative_communication_table}
	\begin{tabular}{@{}ccc@{}}
		\multirow{2}{*}{} & \multicolumn{2}{c}{Cooperative Communication} \\ \cmidrule(l){2-3} 
		& agent1                	& agent2                \\ \cmidrule(l){2-3} 
		MADDPG     &  \textbf{1.388906879}	& \textbf{1.388906879} 	\\ \cmidrule(l){1-3} 
		RMADDPG   &  0.796367905			& 0.796367905 			\\ \cmidrule(l){1-3} 
		T1-HRTMADDPG  &  0.709515411			& 0.709515411 			\\ \cmidrule(l){1-3} 
		T2-HRTMADDPG  &  1.043408619			& 1.043408619 			\\ \cmidrule(l){1-3} 
		T3-HRTMADDPG  &  0.280895594			& 0.280895594 			\\ \cmidrule(l){1-3} 
		T4-HRTMADDPG  &  -0.274622019			& -0.274622019 			\\ \cmidrule(l){1-3} 
		T5-HRTMADDPG  &  -0.005085861			& -0.005085861			\\ \cmidrule(l){1-3} 
	\end{tabular}
\end{table}
	
\subsubsection{Predator-prey}
	
There are also three agents in this environment, two of which are Predator and the other one is Prey with a fast speed.
Multiple obstacles are placed in this environment.Predator is trying to collide with Prey to get the reward, and Prey is trying to keep the predator away from the target.
The reward function for each Predator is:
\begin{equation} 
R = \sum_{t=1}^{}d(p,prey_i)
\end{equation}

On the other hand, the reward function for each Prey is:
\begin{equation} 
R = \sum_{t=1}^{}d(p,predator_i)
\end{equation}

\begin{figure}[!h]
	\centerline{\includegraphics[width=4in]{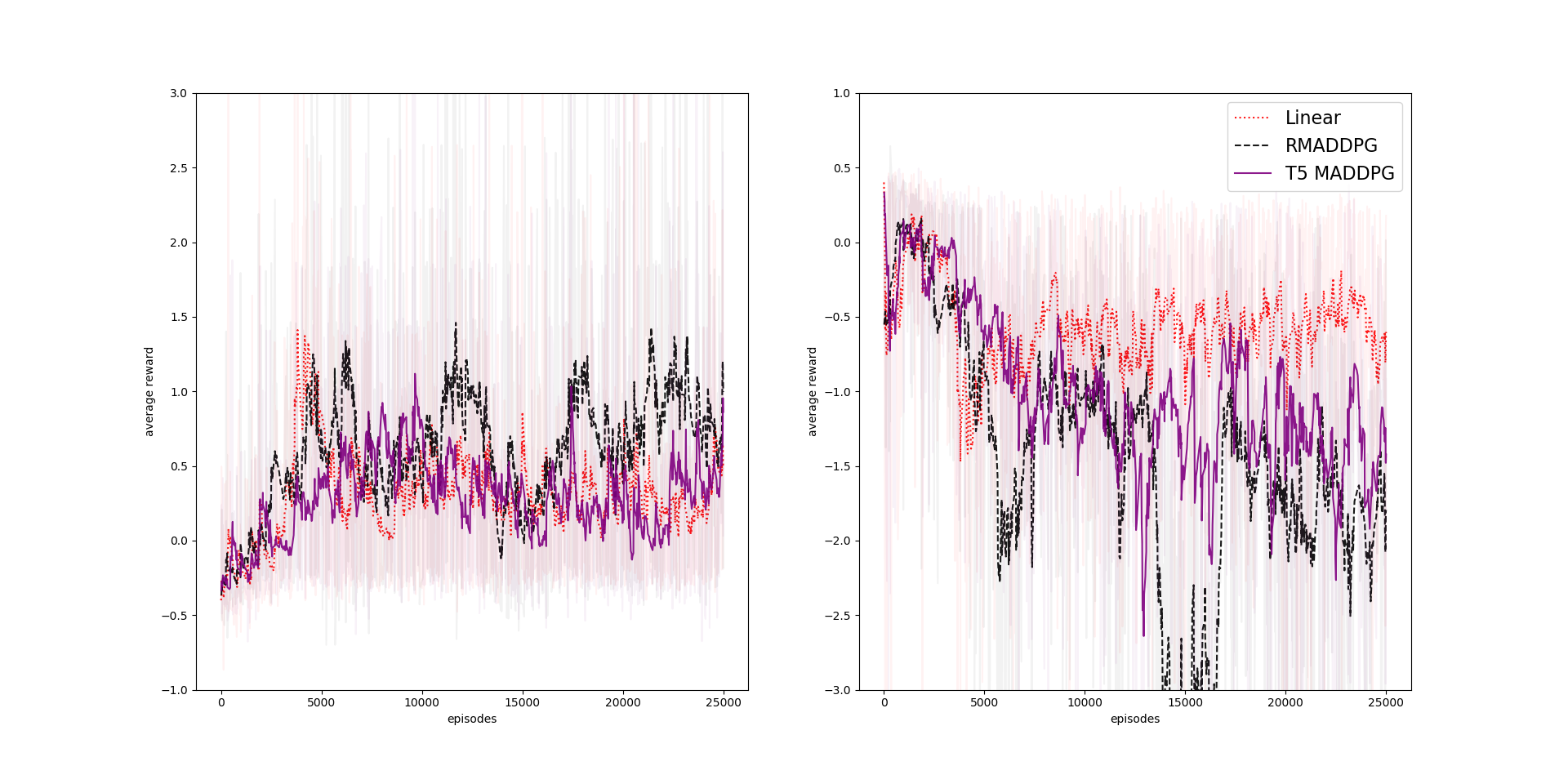}}
	\caption{Predator prey}
	\label{predator_prey}
\end{figure}

As shown in Figure \ref{predator_prey} and Table \ref{predator_prey_table}
In the experimental Predat-prey environment, with the deepening of Transformer Encoding level,
The average reward value of HRTMADDPG training and the average reward value of test are both increasing and gradually approaching the performance of RMADDPG.
However, both RMADDPG and HRTMADDPG fluctuated greatly in the average reward value under this environment.
In the test environment, under the I1-HRTMADDPG algorithm, the average reward value of the three Predator agents is large.
Prey has a larger average reward value under MADDPG algorithm.

\begin{table}[!h]
	\centering
	\caption{Reward of the proposed DRL framework at predator prey for each agent}
	\label{predator_prey_table}
	\begin{tabular}{@{}ccccl@{}}
		\multirow{2}{*}{} & \multicolumn{4}{c}{Predator-Prey} \\ \cmidrule(l){2-5} 
		& agent1  			& agent2 			& agent3 			& agent4 				\\ \cmidrule(lr){2-4}
		MADDPG    & 0.2440855			&  0.2440855		& 0.2440855			& \textbf{-0.5082402}	\\ \cmidrule(lr){1-5}
		RMADDPG  & 0.8724153 			&  0.8724153		& 0.8724153 		& -2.07498				\\ \cmidrule(lr){1-5}
		T1-HRTMADDPG & \textbf{1.36596}	&  \textbf{1.36596}	& \textbf{1.36596}	& -1.653192				\\ \cmidrule(lr){1-5}
		T2-HRTMADDPG & 1.195775			&  1.195775			& 1.195775			& -1.723919				\\ \cmidrule(lr){1-5}
		T3-HRTMADDPG & 0.813983			&  0.813983			& 0.813983			& -1.472489				\\ \cmidrule(lr){1-5}
		T4-HRTMADDPG & 0.5922874			&  0.5922874		& 0.5922874			& -1.444076				\\ \cmidrule(lr){1-5}
		T5-HRTMADDPG & 0.9802052			&  0.9802052		& 0.9802052			& -1.571625				\\ \cmidrule(lr){1-5}
	\end{tabular}
\end{table}

\subsection{Compare with R-MADDPG}
\begin{figure}[!h]
	\centerline{\includegraphics[width=4in]{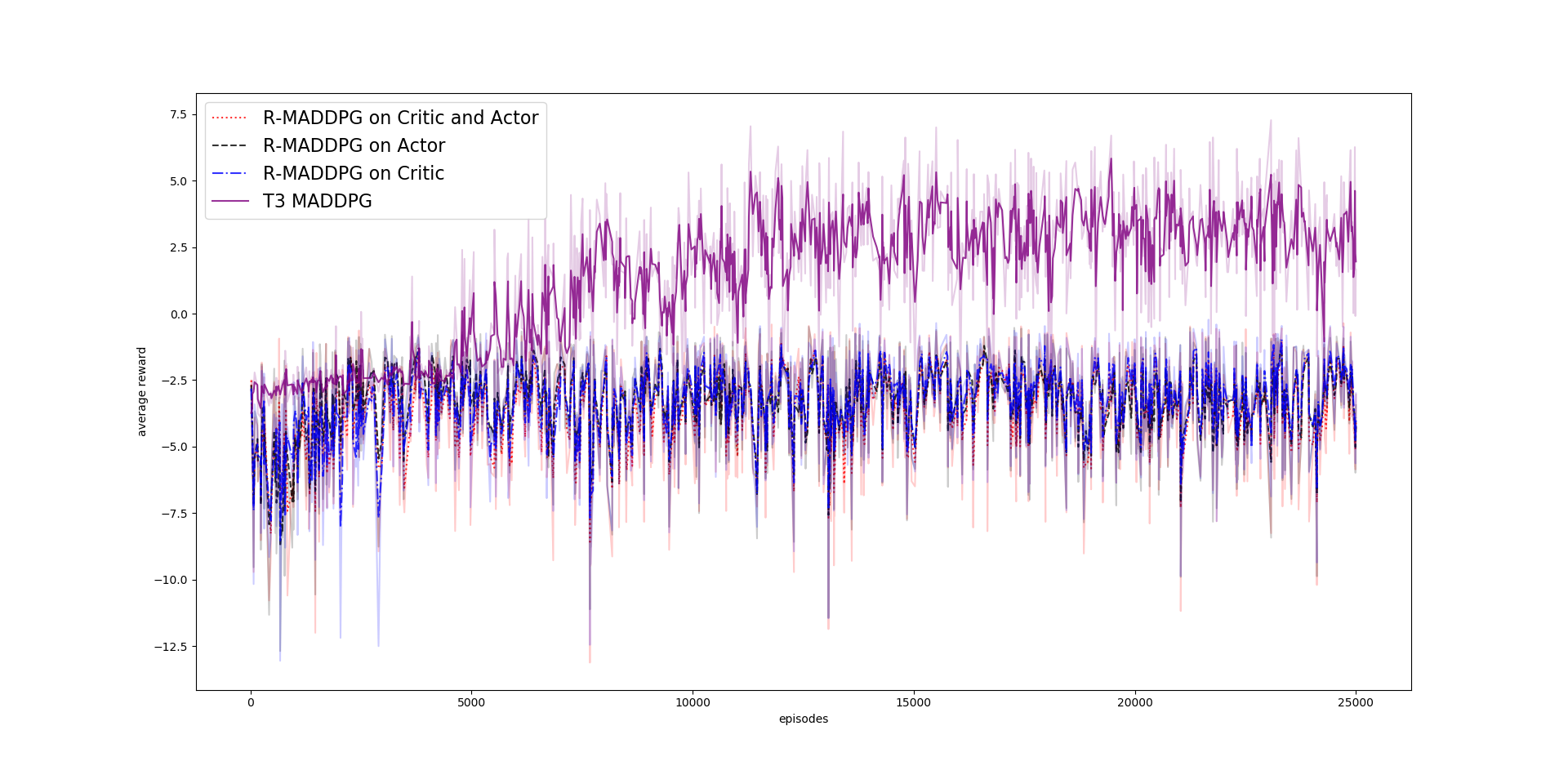}}
	\caption{Compare with R-MADDPG}
	\label{comaprermaddpg}
\end{figure}

The recently-proposed R-MADDPG\cite{wang2020rmaddpg} uses LSTM based on the MADDPG have achieved a great deal in partially-observable environments. 
she used recurrent actor-critic networks, their recurrent layers had the purpose of remembering the messages from previous time steps to solve certain domains that are not allowed to communicate every time step. 
Combined with the above experiments, the method we proposed performs best in a Cooperative navigation environment.
To verify the accuracy of our experiment, we compare these results of R-MADDPG on Critic and Action, R-MADDPG on Critic, R-MADDPG on Actor and our proposed method HRTMADDPG.
As shown in Figure \ref{comaprermaddpg}, the speed of reward curve fitting is greatly improved with the deepening of Transformer Encoding level.
The average HRTMADDPG reward is growing faster than R-MADDPG.

\section{CONCLUSIONS}

In this work, we focus on the two methods of MARL and the Transformers network, we propose a joint MARL framework, Hierarchical RNNs-Based Transformers MADDPG(HRTMADDPG). 
Transformer has demonstrated superior performance in capturing the correlations between multiple time sequences. 
Moreover, we proposed ways to encode two explicit factors in MARL that can capture the causality between sequences and make HRTMADDPG more efficient. The experimental results on a benchmark MADDPG corpus verified the effectiveness and superiority of our approach.
The results demonstrate that the RTHMARDPG is more efficient than the MADDPG and RMADDPG. 
However, there are still great defects in the experiment. 
Regarding their time performance, 
there is still some work to do in the future. 
For example, 
Experiments show that our algorithm has a good effect on cooperative navigation, but not on multiagents cooperation or multiagents confrontation.
Furthermore, how to design and fine-tune the reward function is necessary; otherwise, it may yield unsatisfactory performance.

\section*{Acknowledgment}

This research is supported by the National Key Research and Development Program of China (Grant No. 2019YFB1406201), National Natural Science Foundation of China(Grant No. 62071434), and the Fundamental Research Funds for the Central Universities(Grant No. CUC210B017).

\ifCLASSOPTIONcaptionsoff
  \newpage
\fi

\renewcommand\refname{Reference}
\bibliography{t-maddpg}

\end{document}